\documentclass[preprint,12pt]{elsarticle}

\usepackage{graphicx}
\usepackage{xcolor}
\usepackage{amssymb}
\usepackage{amsmath}
\usepackage{lineno}

\usepackage{booktabs}
\usepackage{listings}

\newcommand{\etal}{\emph{et al.}}
\newcommand{\smallsectionstart}[1]{\noindent\textbf{#1}}

\lstdefinelanguage{pseudo}{
    morestring=[b]',
    keywords = {function,procedure,end,if,else,repeat,until,foreach,return,break,or,and,while,pick},
    numbers=left,
    comment=[l]{//}
}
\lstdefinestyle{pseudocodeStyle}{basicstyle=\footnotesize,breaklines=true,language=pseudo}

\journal{Journal Name}

\begin{document}

\begin{frontmatter}

\title{A Multi-Objective Anytime Rule Mining System to Ease Iterative Feedback from Domain Experts}

\author{Tobias Baum}
\ead{tobias.baum@inf.uni-hannover.de}
\author{Steffen Herbold}
\ead{herbold@cs.uni-goettingen.de}
\author{Kurt Schneider}
\ead{kurt.schneider@inf.uni-hannover.de}

\begin{abstract}
Data extracted from software repositories is used intensively in Software Engineering research, for example, to predict defects in source code. In our research in this area, with data from open source projects as well as an industrial partner, we noticed several shortcomings of conventional data mining approaches for classification problems:
(1)~Domain experts' acceptance is of critical importance, and domain experts can provide valuable input, but it is hard to use this feedback.
(2)~The evaluation of the model is not a simple matter of calculating AUC or accuracy. Instead, there are multiple objectives of varying importance, but their importance cannot be easily quantified. Furthermore, the performance of the model cannot be evaluated on a per-instance level in our case, because it shares aspects with the set cover problem.
To overcome these problems, we take a holistic approach and develop a rule mining system that simplifies iterative feedback from domain experts and can easily incorporate the domain-specific evaluation needs. A central part of the system is a novel multi-objective anytime rule mining algorithm. The algorithm is based on the GRASP-PR meta-heuristic but extends it with ideas from several other approaches.
We successfully applied the system in the industrial context. In the current article, we focus on the description of the algorithm and the concepts of the system. We provide an implementation of the system for reuse.
\end{abstract}


\end{frontmatter}

\section{Introduction}

This report introduces a rule mining system that overcomes several shortcomings we perceive in commonly used data mining approaches. We developed the system for a study to mine data on source code reviews gathered from software development repositories~\cite{baum2018miningStudy}. The system's design is based on the belief that currently, the most promising approach to extract knowledge from data is to let human and computer work as a team~\cite{holzinger2016interactive,ankerst2002report}. Both can bring their strengths into the combination: The computer can sift through vast amounts of data swiftly and without loss of concentration. Human domain experts can provide input that is not readily available from the data, for example on preferences or on conditions that hold in the respective context. Like every design process, this teamwork benefits from being iterative: The humans can see preliminary results from the computer and provide focus or new insights based on the results.

Current approaches for data mining fall short from this ideal of iterative collaboration in several ways:
\begin{itemize}
\item Some mining approaches create opaque models that cannot be analyzed by domain experts at all. Some types of models (e.g., support vector machines or neural networks) are nearly always hard to understand, while for others (e.g., rulesets or decision trees) understandability depends on complexity~\cite{dam2018explainable}.
\item It is often hard to map feedback from domain experts to the parameters needed by the mining algorithm.
\item One type of human input concerns the cost of misclassification. Many approaches are not cost-sensitive at all or need a cost matrix to be specified at the start. In reality, problems are cost-sensitive, but the exact cost matrix is often not known.
\item Most approaches allow input to be given only at the start of a run and create a single model as the result of such a run. This limits the points in time when domain experts can give feedback.
\item The knowledge discovery process encompasses multiple phases~\cite{mariscal2010survey}, including cleaning of the data, selection or perhaps even creation of features, and interpretation of the results. Some systems, like Weka, combine support for many of these steps, but in a very general and not streamlined way.
\end{itemize}

In the current report, we propose a data mining \emph{system} to overcome these shortcomings. An essential part of this system is a rule mining algorithm, but it also contains other features to ease iterative collaboration between the system and human domain experts. Our system helps to overcome the above-mentioned limitations because it is
\begin{itemize}
\item Multi-objective: The system can find rules with varying trade-offs for objectives like reduction of false positives, reduction of false negatives and simplicity. This avoids the need to specify a cost matrix at the start.
\item Interactive: The user can interactively explore the data and the results and provide feedback to guide the mining.
\item Iterative: The feedback by the user is integrated into the mining process and can be iteratively refined. This also includes the possibility to undo earlier decisions.
\item Anytime: The user can explore the data and the current results and interact with the system at any time. The user rarely needs to wait for the system, and neither does the system need to wait for the user.
\item Geared towards domain expert feedback: The user can provide feedback without the need to translate it into opaque tuning parameters.
\end{itemize}

In the next section, we provide background information on our application area and motivate and formalize the choice of a rule mining approach. We then go on to describe the rule mining algorithm, still largely ignoring interactive features (Section~\ref{sec:algorithm}). After a short detour into possibilities to adopt our concepts in a domain-specific way (Section~\ref{sec:domainSpecific}), we describe the interactive features of the system (Sections \ref{sec:interactiveFeedback}~and~\ref{sec:interactiveExploration}). Finally, we deal with the problem of scientific rigor, mainly reproducibility, for a system with the human in-the-loop (Section~\ref{sec:scientificRigor}).

\section{Motivation and Model Choice}

\subsection{Application Context: Review Remark Prediction}
\label{sec:applicationContextRemarkPrediciton}

The motivation for the data mining system we present in this article stems from our software engineering research on code reviews. Code reviews are a software quality assurance technique in which code or code changes are manually checked by one or more developers~\cite{baum2016qrs}. In their modern change-based form, they are widely used in industrial practice~\cite{rigby2012contemporary,baum2017profes}. When a reviewer spots a defect or some other point that needs to be discussed, he or she usually creates a `review remark' that will be communicated to the code's author. In our research on code reviews, we are interested in finding parts of the code change that will not lead to review remarks. These parts can then be left out from the review, saving time and leaving more mental resources for the rest of the review. The underlying problem behind some review remarks can be manifested in several parts of the code, and it suffices to check one of them to note it.
In a case study in a medium-sized software company, we extracted data on which code change parts led to which remarks from software repositories~\cite{baum2018miningStudy}. The rule mining system described in the current article was then used to derive rules that characterize change parts that do not need to be reviewed. The extracted data spanned about five years of the company's development history and consisted of 730,674 records/instances, linked to 70,900 review remarks.

\subsection{Problem Statement}
\label{sec:problemStatement}

\begin{figure}
    \centering
    \includegraphics[width=0.7\textwidth]{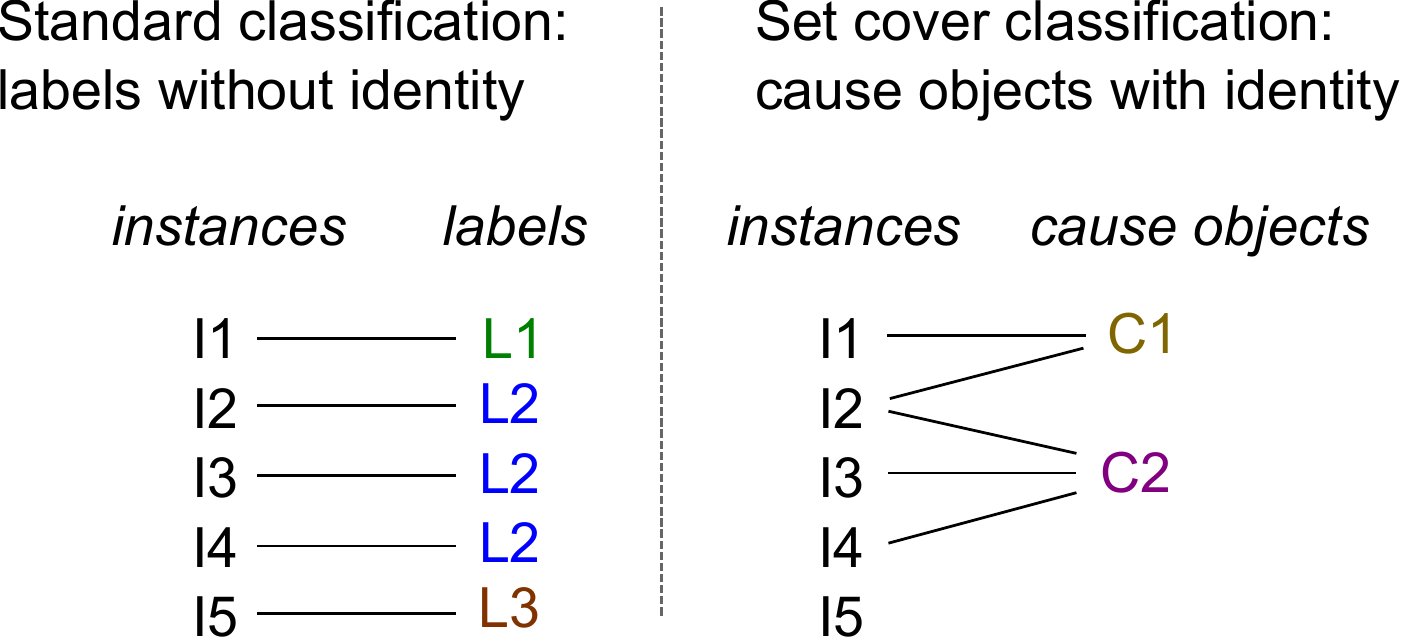}
    \caption{Set Cover Classification versus Standard Classification}
    \label{fig:setCoverExample}
\end{figure}

We will now state the mining problem more abstractly:
Consider a problem, like the review remark prediction problem introduced above, in which an underlying cause (e.g., problem behind a review remark) is represented in several instances. One instance can contain representations of one, several or no causes. The goal is to minimize some effort to treat the underlying causes. We want to solve this problem by predicting which instances to select to spend the effort on. Figure~\ref{fig:setCoverExample} illustrates such a situation: To cover C1 and C2, it is sufficient to select I2. Other sufficient, but sub-optimal, selections would be I1 and I3 or I1 and I4. We call selecting an instance a \emph{positive} prediction and not selecting it a \emph{negative} prediction. Moreover, we assume that there might be multiple instances where spending the effort would lead to treating the same cause, as in the example above. To minimize the effort, we only want to spend the effort on a single of these instances, instead of all of them. Furthermore, we assume that domain experts who are responsible for the treatments do not simply accept predictions, but want to understand the reasoning and be able to modify it using their domain knowledge. Otherwise, they will not use the predictive model. Thus, the problem we want to solve has the following properties:
\begin{itemize}
\item Minimizes the number of positive predictions (i.e., predicted instances) while still covering (nearly) all causes.
\item Account for the relationship between causes and instances.
\item Interpretable and modifiable by experts.
\end{itemize}

In general, this is a classification problem. There are many ways to describe hypotheses for the classification of objects, e.g., through support vector machines, neural networks, regression functions like logistic regression, and rules. Of these approaches, rules have the advantage that they are easy to interpret and modify by humans. Within this paper, we consider rules described as boolean expressions with the expressiveness of a disjunctive normal form, i.e.,
\begin{equation}
    C_1 \vee ... \vee  C_p
\end{equation}
where $C_k = c_{k,1} \wedge ... \wedge c_{k,l_k}$ are conjunctions over atomic conditions. Every boolean expression over can also be described as a DNF. Another popular way to describe rules through boolean expressions is decision trees~\cite{dataMining2011}. Decision trees have equal expressive power to DNFs. While it is easier for humans to use decision trees when evaluating them for a specific data point~\cite{huysmans2011empirical}, we believe that rulesets are preferable for understanding. In an unordered ruleset, each rule can be treated as an independent nugget of knowledge and clearly belongs to a certain class. In contrast, a node in a decision tree has to be interpreted with all previous nodes in mind, and the class is only known when reaching a leaf node. So we chose rulesets as the representation of knowledge in the proposed system.

Given a training set with instances $X=\{x_1, ..., x_m\} \subset \mathbb{R}^{n}$ with boolean labels $Y = \{y_1, ..., y_m\} \in \{0,1\}^m$ rule mining algorithms try to find a DNF that postulates logical rules of the instances $x_i \in X$ of the form $x_{i,j} \leq A$, $x_{i,j} \geq A$, $x_{i,j} = A$, or $x_{i,j} \neq A$ for a constant $A \in \mathbb{R}$. Thus, such rule mining algorithms determine a hypothesis 
\begin{equation}
    h: \mathbb{R}^n \to \{0,1\}
\end{equation}
such that $h$ is a DNF of atomic conditions over the real-valued vectors. The problem with learning DNFs is that the search space is huge. Thus, a good strategy to search for DNFs that yield good results is required. One of the most popular algorithms for mining such rules is the RIPPER algorithm~\cite{Cohen1995}, which uses a greedy strategy for the inference of the rules. The algorithm first greedily creates rules: new atomic conditions over features are added by using information gain as the criterion for the feature selection until no negative examples are covered. This process leads to overly complex rules that overfit the data. To counter overfitting, the rules are pruned such that the ratio $\frac{tp-fp}{tp+fp}$ is optimized. This algorithm efficiently searches for good rules given that positive and negative examples are equally important for the use case. While it is certainly possible to use RIPPER for our problem, several limitations are likely to degrade the performance for the prediction task. We discuss these limitations of RIPPER in the following and use this to outline the requirements on an algorithm to solve our problem in greater detail. We note that while we discuss our problem for RIPPER, the problems for other approaches, e.g., based on decision tree learning (C4.5, PART, CART) would be similar. 

\subsection{Rule Set Syntax}

In the preceding subsection, we formally introduced our problem and motivated the use of rulesets that are expressible as DNF. Normal forms can help human understanding by ensuring a structured representation. However, this representation can be inefficient, needing a complex ruleset for a simple concept. Based on the intuition that it might be easier to express the opposite concept in such a case, we use a combination of two formulae in disjunctive normal form:

\begin{lstlisting}
   (... and ... and ...)
   or (... and ... and ...)
   ...
except
   (... and ... and ...)
   or (...)
   ...
\end{lstlisting}
Or, put mathematically: $\bigvee_{r \in incl}\left(\bigwedge_{c \in r} c \right) \wedge \neg \bigvee_{r \in excl}\left(\bigwedge_{c \in r} c \right)$ (with ``incl'' being the inclusions, i.e., the rules before the ``except'', and ``excl'' being the exclusions, i.e., the rules after the ``except''). This notion allows certain rules to be written more concisely than a plain disjunctive normal form but is still simple and easily explainable. Furthermore, every single rule in the disjunctions can be treated as a separate nugget of knowledge.

In the following, we use this vocabulary:
\begin{description}
\item[Ruleset] The whole model (as shown in the example above).
\item[Rule] A single disjunction in the model.
\item[Proposition] One part of the conjunction in a rule. For nominal features, this can be a comparison with ``equals'' or ``not equals'', for numeric features it can be a comparison with ``less or equal'' or ``greater or equal''.
\end{description}

\subsection{Multi-objective rule learning}

For domain-specific use cases, there is a cost associated both with false positive predictions and false negative predictions. These costs can be modeled in a cost function $cost(h,X,Y)$ that estimates the costs using labeled data. For the remainder of this paper, we assume without loss of generality that cost functions should be minimized for optimal performance. Thus, a learning algorithm should ideally directly optimize this cost function, i.e., 
\begin{equation}
    \min_h cost(h,X,Y).
\end{equation} 
In general, use cases might have multiple competing cost objectives, e.g., minimizing the time to market and minimizing the costs of the development. This leads to a multi-objective optimization problem of the form
\begin{equation}
    \min_h (cost_1(h,X,Y), ..., cost_o(h,X,Y))
\end{equation}

Usually, there is no single solution which is optimal for all cost functions. Therefore, we consider Pareto optimal solutions instead, i.e., solutions that are not dominated by any other solution. One solution $h$ dominates another solution $h'$ if $cost(h,X,Y)\leq cost(h',X,Y)$ for all $cost \in {cost_1, ..., cost_o}$ and there exists at least one cost function $cost\in {cost_1, ..., cost_o}$ such that $cost(h,X,Y) < cost(h',X,Y)$. While it is certainly possible to use RIPPER and afterward calculate costs functions, the algorithm itself ignores the cost functions during the training. It would be possible to modify the growing and pruning criteria to account for a single cost function directly. However, RIPPER cannot easily be modified to determine Pareto optimal solutions in a multi-objective setting. For our use case, we have three types of cost functions: 1) cost functions for the estimation of the effort spent, 2) cost functions for the number of causes covered, and 3) costs for the cognitive complexity required for the understanding of rules.

\subsection{Relationships between instances}

Another property of RIPPER is that the label of an instance only depends on the instance itself, not on other instances. Given that the instances are independent given the class label, this is not a problem. However, this is not the case for our type of problem. Assume that a subset $\bar{X} \subset X$ of instances all have the same reason for the positive label. In order to identify the cause, it is sufficient to identify one instance of $\bar{X}$. Optimizing for both the effort and the number of causes covered, this relationship must be considered. There is no way to modify RIPPER to account for this relationship without completely changing the algorithm. 

\subsection{Feedback}

Finally, for our problem domain experts must not only be able to understand the generated hypothesis but modify it with feedback. While they certainly could modify DNFs produced by RIPPER, the implications of these modifications for the performance would be unclear. To solve the problem, the learning should directly interact with domain experts and allow for the following: 
\begin{itemize}
    \item Definition of rules by domain experts that must be part of the hypothesis
    \item Definition of atomic conditions or rule patterns that must not be part of the hypothesis
    \item Analysis of the impact of manual modifications of the hypothesis concerning the Pareto front 
\end{itemize}

This interaction should be iterative and allow users to modify the manually defined conditions and get feedback about the results at any time.

\section{The Rule Mining Algorithm}
\label{sec:algorithm}

\begin{figure}
    \centering
    \includegraphics[width=0.8\textwidth]{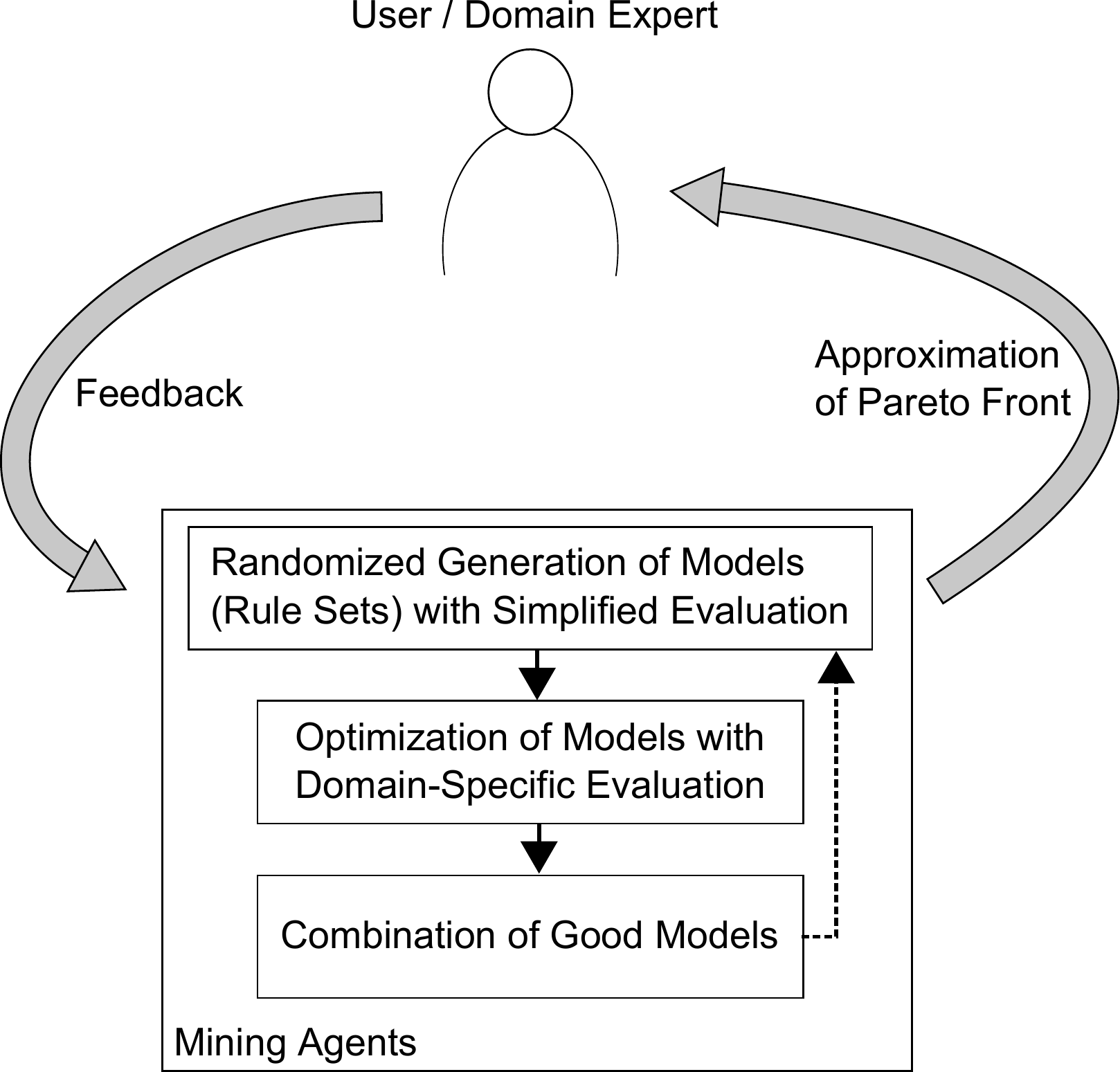}
    \caption{Schematic Overview of the Mining Approach}
    \label{fig:approachOverview}
\end{figure}

\noindent The algorithm combines ideas from:
\begin{itemize}
\item The GRASP-PR meta-heuristic~\cite{resende2016optimization}
\item Multi-Objective Evolutionary Algorithms (MOEAs), especially MSOPS~\cite{wagner2007pareto}
\item Separate-and-conquer rule learning~\cite{janssen2010quest,furnkranz2015brief}
\item Random Forests~\cite{breiman2001random}
\item Constraint-based data mining~\cite{han1999constraint}
\end{itemize}

\noindent Like in GRASP-PR, the algorithm consists of three main parts:
\begin{enumerate}
\item greedy, randomized generation of candidate solutions (rulesets)
\item optimization of rulesets by local search
\item combination (path relinking) of rulesets for further improvement
\end{enumerate}

\begin{figure}[tbp]
\centering
\begin{lstlisting}[style=pseudocodeStyle]
repeat
  if there is work on the local search queue
    ruleToWorkOn := blackboard.takeFromLocalSearchQueue()
    localSearch(
      combineRuleSets(ruleToWorkOn, blackboard.getBestResult()), 
      blackboard.currentTargetFunction)
    resultParetoSet := localSearch(
      ruleToWorkOn,
      blackboard.currentTargetFunction)
    blackboard.addToPathRelinkingQueue(best item from resultParetoSet)
    blackboard.addToPathRelinkingQueue(random item from resultParetoSet)
  else if there is work on the path relinking queue
    ruleToWorkOn := blackboard.takeFromPathRelinkingQueue()
    pathRelink(
        ruleToWorkOn, 
        blackboard.getBestResult(), 
        blackboard.currentTargetFunction)
    pathRelink(
        ruleToWorkOn,
        blackboard.getRandomResult(),
        blackboard.currentTargetFunction)
  else
    pick one at random:
      //perfection by path relinking
      pathRelink(
          blackboard.getBestResult(),
          blackboard.getRandomResult(),
          blackboard.currentTargetFunction)
    or
      //perfection by local search
      localSearch(
          blackboard.getRandomResult(),
          blackboard.currentTargetFunction)
    or
      //generation of new rule material
      newRuleSet := generateNewRuleSet(iteration count)
      blackboard.addToParetoFrontIfPossible(newRuleSet)
      blackboard.addToLocalSearchQueue(newRuleSet)
    end-pick
  end-if
until the user stops the agent
\end{lstlisting}
\caption{Pseudo code for the agent's main loop (simplified; see~\cite{baum2018predictionOnlineMaterials} for full source code)}
\label{fig:agentMainLoop}
\end{figure}

An overview of the approach can be seen in Figure~\ref{fig:approachOverview}.
Unlike in standard GRASP-PR, the steps are executed by one or several \emph{mining agents} that can run in parallel. Each agent runs in an infinite loop in which it takes a work package, e.g., a starting rule that needs to be optimized by local search, from a queue of work items. When there is no open work, it generates new items, or it tries to improve existing solutions further. The pseudo-code for the agents' main loop is shown in Figure~\ref{fig:agentMainLoop}. A second difference to standard GRASP-PR is that our algorithm works on a Pareto front of solutions and takes the multi-objective nature of the problem into account.
For parts of the algorithm that need an absolute order between objective vectors instead of the partial Pareto order, there is a `current target function' that combines a cost vector into a single number. It will be used again in the later sections.

Because it is conceptually similar to the blackboard in blackboard algorithms~\cite{buschmann1996patterns}, the data structure that keeps the current Pareto front and the queues with open work items will be called ``blackboard'' in the following. Having the mining agents run in parallel to each other and to the user on a shared blackboard allows the user to see the current state of the search at any time. The user can explore the best rules found so far, without the need to wait for a mining job to complete. This is the foundation for interactive feedback from the user to the mining agents, which will be discussed in Section~\ref{sec:interactiveFeedback}.

\subsection{Rule Set Generation}

\begin{figure}[tbp]
\centering
\begin{lstlisting}[style=pseudocodeStyle]
function generateNewRuleSet(countLimit)
  data := sampleSubsetOfFeatures(blackboard.allData)
  data := sampleSubsetOfRecords(data)
  
  ruleSet := empty ruleset
  
  maxExclusionRuleCount := random number between 0 .. countLimit
  uncovered := data
  while there are less than maxExclusionRuleCount exclusion rules
    searchBias := select random search bias
    rule := findExclusionRuleByGreedyTopDown(uncovered, searchBias)
    if no rule found
      break
    end-if
    ruleSet := ruleSet.addExclusion(rule)
    uncovered := removeCoveredRecords(uncovered, rule)
  end-while
  
  maxInclusionRuleCount := random number between 0 .. countLimit
  uncovered := data  
  while there are less than maxInclusionRuleCount inclusion rules
    searchBias := select random search bias
    rule := findInclusionRuleByGreedyTopDown(uncovered, searchBias)
    if no rule found
      break
    end-if
    ruleSet := ruleSet.addInclusion(rule)
    uncovered := removeCoveredRecords(uncovered, rule)
  end-while
  
  return ruleSet
end-function
\end{lstlisting}
\caption{Pseudo code for the generation of new rulesets (simplified; see~\cite{baum2018predictionOnlineMaterials} for full source code)}
\label{fig:ruleSetGeneration}
\end{figure}

One key component of GRASP heuristics is a randomized, greedy heuristic to generate new solutions that will then be optimized further. In this subsection, we describe the corresponding part of our mining algorithm. It is based on a standard greedy separate-and-conquer rule learning algorithm. The algorithm is randomized in several ways:
\begin{itemize}
\item A random subset of features is selected (random subspace method~\cite{ho1998random}).
\item A random subset of records is selected. Additionally, this sub-sampling down-samples the majority class.
\item The maximum number of rules in the ruleset is chosen at random.
\item The search bias for the greedy top-down rule search is chosen randomly. The bias is selected based on search biases found to be promising in the literature~\cite{janssen2010quest}: Precision, Laplace, relative cost and m-estimate. There is also a small probability of choosing a random rule. That means that given infinite time, the algorithm will cover the full rule space.
\end{itemize}
Figure~\ref{fig:ruleSetGeneration} outlines the basic algorithm. 
The randomization is similar to random forests~\cite{breiman2001random}\footnote{obviously with rules instead of decision trees}, but instead of combining the singular models into an aggregate model, the models are used to improve the Pareto front and as a seed for the further optimization steps.

The inclusion and exclusion parts of the ruleset are generated independently. It is up to the optimization steps to remove redundant rules created this way.

\subsection{Local Search}

\begin{figure}[tbp]
\centering
\begin{lstlisting}[style=pseudocodeStyle]
function localSearch(initialRuleSet, targetFunction)
  candidateRules := all rules from initialRuleSet
  neighborhoodType := RULE_ADDING
  resultParetoSet := {initialRuleSet}
  resultParetoSet.add(emptyRuleSet())
  current := emptyRuleSet()
  
  repeat
    possibleMoves := rulesets in neighborhood of current ruleset, restricted to neighborhoodType
    shuffle possibleMoves randomly
    
    //add neighbors to Pareto set and determine best neighbor for targetFunction
    bestNeighbor := current
    foreach move in possibleMoves
      resultParetoSet.add(move)
      if (targetFunction(move) < targetFunction(bestNeighbor)
          or (targetFunction(move) == targetFunction(bestNeighbor) and move could be added))
        bestNeighbor := move
        isOnPlateau := objectiveVector(current) == objectiveVector(bestNeighbor)
      end-if
    end-foreach
    
    if bestNeighbor == current
        or (isOnPlateau and search has been on plateau for too long)
      //no improvement found, leave the current neighborhood
      if neighborhoodType == RULE_ADJUSTING
        neighborhoodType := RULE_ADDING
      else
        ruleAddingDidNotImproveAnything := true
      end-if
    else
      //move to best neighbor and try to improve it by adjusting
      current := bestNeighbor
      neighborhoodType := RULE_ADJUSTING
    end-if
  until ruleAddingDidNotImproveAnything
  return resultParetoSet
end-function
\end{lstlisting}
\caption{Pseudo code for local search (simplified; see~\cite{baum2018predictionOnlineMaterials} for full source code)}
\label{fig:localSearch}
\end{figure}

The basic structure of the local search is that of a hill-climbing meta-heuristic. It was adapted to make it suitable for a multi-objective cost function. The local search keeps a Pareto set of visited rules, which forms the result at the end of the search. The search itself is guided by a target function that collapses the objective vector into a single cost value (i.e., smaller values are better). In each iteration, the algorithm moves to the neighbor with the smallest target function value. When the search is on a plateau, i.e., the target function value stays the same in the best case, it moves to one of these neighbors with equal cost value. Such a `plateau move' is only performed a limited number of times and only to neighbors that could at least be added to the Pareto set. In this case, the Pareto set has a role similar to the tabu list in tabu search~\cite{glover1989tabu} and ensures that once visited solutions will not be revisited. A pseudo-code version of the algorithm can be seen in Figure~\ref{fig:localSearch}.

To reduce the size of the neighborhood and therefore reduce the time needed for the search, the search alternates between two restricted types of neighborhood: In the `rule adding' neighborhood, the neighbors are determined by adding a candidate rule to the current ruleset. In the `rule adjusting' neighborhood, the rule that was added last to the ruleset is adjusted: A proposition is removed, the value of a numeric comparison is changed to the nearest split point, or a random proposition is added. Search in the `rule adjusting' neighborhood is kept up as long as an improvement is found, then `rule adding' is tried for one iteration again. It might be a bit surprising that the search does not start with the input rule, but instead works forward from the empty rule. This is a performance optimization to allow better caching of evaluated rulesets, and also increases the algorithm's bias towards simple rulesets.

\subsection{Path Relinking}

Often, the singular rules in a ruleset are mostly independent of each other. Therefore, it is highly likely that combining two good rulesets will lead to another good ruleset. In the GRASP literature, this process of combining solutions is called path relinking. It begins by determining the difference between two rulesets, the start and end ruleset. The difference can be regarded as a collection of independent actions like `add rule X' or `remove rule Y' that move the start rule closer to the end rule. Beginning at the start rule, the algorithm looks for a good action to apply next, applies it, and iterates until the end rule is reached. In this process, every visited candidate is compared to the Pareto front and added if possible. Pseudo code for the algorithm can be seen in Figure~\ref{fig:pathRelinking}.

\begin{figure}[tbp]
\centering
\begin{lstlisting}[style=pseudocodeStyle]
procedure pathRelink(startRuleSet, endRuleSet, targetFunction)
  relinkActions := differences from startRuleSet to endRuleSet
  current := startRuleSet
  while relinkActions is not empty
    goodAction := determineGoodAction(current, relinkActions)
    current := goodAction.apply(current)
    relinkActions.remove(goodAction)
  end-while
end-procedure

function determineGoodAction(current, relinkActions, targetFunction)
  currentCost := targetFunction(current)
  bestValue := positive infinity
  bestAction := null
  foreach action in relinkActions
    candidate := action.apply(current)
    candidateValue := targetFunction(candidate)
    //implicitly add each candidate to Pareto front if possible
    try to add candidate to Pareto front
    if candidateValue < currentCost
      //stop at the first improvement to keep the number of evaluations down
      return action
    end-if
    
    if (candidateValue < bestValue or (candidateValue == bestValue and candidate dominates bestAction.apply(current)))
      bestValue := targetFunction(candidate)
      bestAction := action
    end-if
  end-foreach
  
  return bestAction
end-function
\end{lstlisting}
\caption{Pseudo code for path relinking (simplified; see~\cite{baum2018predictionOnlineMaterials} for full source code)}
\label{fig:pathRelinking}
\end{figure}

Like for other parts of the algorithm, the decision what a `good action' is depends on a target function that collapses the objective vector into a single cost value. To keep the number of full evaluations down, we do not look for the best action, but stop the search as soon as we find one that improves upon the start value. To make this work best, the start rule should already have a good target function value. When none of the actions leads to an absolute improvement, the algorithm picks the least bad action and uses Pareto dominance when breaking ties for equal target function values.

\section{Integration of the Domain-Specific Aspects of the Problem}
\label{sec:domainSpecific}

Creation of a data mining system is not an end in itself, its a means towards an end. In most of the current paper, we discuss the proposed data mining system without reference to the specific domain that we described in Section~\ref{sec:applicationContextRemarkPrediciton}. The discussion of domain-specific aspects in the current section shows the motivation for some of the design decisions and shows parts of the algorithm that can easily be adjusted to non-standard aspects of a data mining problem.

The most important domain-specific aspect is that our problem was not a simple binary classification problem. Section~\ref{sec:problemStatement} formalizes that every instance is associated with a (potentially empty) collection of cause objects. Our goal was to find an optimal ruleset that covers as many of these objects as possible. This domain-specific aspect could be integrated by two adjustments to the system: (1)~Many parts of the algorithm depend only on a cost function that calculates the cost vector for a ruleset. This cost function can be changed to take the object covering into account. The changed cost function is expensive to compute (about 0.5 seconds per evaluation in our case), so the rest of the system was shaped to avoid excessive cost calculations. (2)~The generation of rules is based on the standard separate-and-conquer rule learning algorithm that demands simple class labels for the records. Such an input can be created from our data by picking a subset of records that cover all objects and labeling these as ``T'' and the others as ``F''. Consistent with the randomized, greedy nature of the rule generation, picking these records is done by a greedy, randomized heuristic for the subset cover problem.

Each record and each coverable object in our case belonged to a part of a source code file in a commit. Therefore, the data can be regarded as hierarchic, e.g., when grouping all records belonging to the same file or commit. This hierarchy was also taken into account in the system, especially in the interactive exploration and data preparation features that will be described in Section~\ref{sec:interactiveExploration}. In this way, we brought the UI closer to the concepts known by domain experts and followed our dictum to ease exploration and feedback by domain experts as much as possible.

\section{Interactive Feedback for the Algorithm}
\label{sec:interactiveFeedback}

In Section~\ref{sec:algorithm}, we described the basic algorithm, but we left out one of the main features: The ability to incorporate interactive feedback from the user. There are four ways how such feedback can be given: By changing the target function for the search, by providing rulesets to consider, by restricting the search space regarding allowed or enforced rules, and by asking for a reduction in the size of the Pareto front. As the mining agents are running in parallel in the background, this feedback can be given at any time.\footnote{To allow the mining agents to keep running when the user switches of the computer, the system is designed to run on a server with a web UI.}

\smallsectionstart{Changing the target function.}
In several parts of the algorithm, the multi-dimensional objective vector needs to be transformed into a single numeric cost value: This single number is used to select the best element when hill-climbing during the local search, as well as during path relinking. It is also used to pick a decent rule as a foundation for further search and generation.
Initially, this target function returns the precision of the ruleset. The user can change it to another function to try out other optimization biases at any time and as often as wanted. A natural extension that has not been implemented in our system so far is to let the user provide a collection of possible target functions. The system can switch at random between these functions so that the search would be broader without repeated user interaction.

\smallsectionstart{Providing rulesets.}
The user might have an idea for a good rule, either from scratch or based on a rule found earlier. He or she can try it out with the tool. Such a user-provided ruleset is treated similarly to a ruleset generated by a mining agent: If it is Pareto-optimal, it is added to the Pareto front. It is also added to the work items for further optimization by local search and path relinking.
To improve responsiveness, there are some differences in the details of the treatment between user-generated and agent-generated rulesets; these can be seen in the system's source code~\cite{baum2018predictionOnlineMaterials}.

\smallsectionstart{Restricting the rule search space.}
A common thought of a user when looking at a generated ruleset might be: ``This rule looks good, but this other rule does not make sense.'' This feedback can be incorporated into the mining process by restricting the search space. One possibility is to reject rules or patterns of rules, for example, all rules with a specific combination of propositions, all rules that compare a numeric feature in a direction deemed invalid, or even all rules using a particular feature. Once the restriction is in place, the generation algorithm will not create such rules anymore, and neither will the local search and path relinking algorithms. Furthermore, all rules that are now forbidden are removed from the Pareto front.

Restricting the search space in the opposite way is also possible: The user can mark a rule as ``accepted''. This will make the rule appear in every ruleset created from now on. Furthermore, the rule is added to all rulesets on the Pareto front, and the combined rulesets are re-evaluated.

It is also possible to undo these restricting actions. To be able to quickly recreate rules that were once found but later declared invalid, invalid rules are cached in the background even after deleting them from the Pareto front.

\smallsectionstart{Reduction of Pareto front size.}
When one or several mining agents are running, they keep on creating new rulesets. After some time, this can lead to a large number of rulesets on the Pareto front. This increases the resource consumption of the system. It can also hinder the interactive exploration of the rulesets, especially if many similar rules are found.
Many multi-objective evolutionary algorithms restrict the number of solutions in the Pareto front by eagerly removing solutions that are close to each other in objective space. Our system does not clean up eagerly but waits for the user to ask for a clean-up. The user also specifies the number of rulesets that shall be kept. By waiting for a user request, the ``knowledge nuggets'' in the mined rules are kept as long as possible. Furthermore, the rulesets to remove are not selected by looking at their distance in objective space. It is easily possible for two semantically different rulesets to have a similar objective vector. Instead, a fingerprint of the rulesets in terms of matched records is created by selecting a random subset of records and recording for each ruleset whether it matches or not. The `distance' between two rulesets is the number of records for which they differ. This distance is used with a standard clustering algorithm to pick one rule per cluster that will be kept, in addition to the optimal rules according to the specified target functions (see above).

\section{Other Features to Allow Fast Iteration}
\label{sec:interactiveExploration}

The feedback provided by the user is based on insights gained by analysis of the data and the found rules. To holistically support fast iteration and feedback, this analysis is supported in several ways by the proposed system.

\subsection{Exploration of the Data}

One possibility for exploration is to analyze aggregate information about subsets of the data. The system can provide statistics for all records matched by a rule or a ruleset, showing aggregates like the most common values, the numeric minimum, maximum, or arithmetic mean for each feature. Similar to the drill-through operation in OLAP\footnote{in a way, the whole data exploration functionality shares similarities with OLAP}, the system provides a sample of the records that match the given criteria. It is also possible to gather a random sample of all records. Such a random sample can be used for further manual analysis, for example for estimating the amount of noise in the classification data. As stated in Section~\ref{sec:domainSpecific}, each record in our use case identified a part of a source code file. By linking from the records to the respective files, the system allowed the user to analyze parts of the data that were not yet explicitly modeled as features in the records.

A typical analysis goal is to analyze parts of the data that are not handled well by the current model. Therefore, it is possible to show all records that are misclassified by a given ruleset. Furthermore, it is also possible to show a sample of records that are not handled explicitly in the model, i.e., that fall into the default branch. Using these features, a domain expert can scrutinize the data and form hypotheses on reasons for the misclassification and on missing features. This process shares many similarities to the `open coding' commonly performed in qualitative data analysis~\cite{glaser1967discovery}.

\subsection{Exploration of the Found Rules}

The user must also be able to explore the output of the system: the Pareto front of found rulesets. This is possible by moving from one ruleset to another along the dimensions of the objective space, either going to neighboring rulesets or directly to the best rule for a dimension. Sometimes, the user is only interested in certain parts of the objective space (e.g., only in rules that are simpler or more accurate than a certain threshold). Therefore, the user can specify a relevant simplex in objective space by specifying bounds for dimensions. These limits are not only used for the interactive exploration of the Pareto front but also focus the mining agents by influencing their selection of the best rule.

When navigating among the found rulesets, it can be tedious to read each rule in detail. Therefore, the user can mark rules as `visited', in addition to the possibility to reject or accept rules (Section~\ref{sec:interactiveFeedback}). Marked rules are highlighted with color in the UI so that the user sees which parts of the ruleset he or she does not need to scrutinize again.


When selecting rulesets as the model for our data mining system, we argued that rules can be understood easily by humans. But the efficiency of understanding likely depends on the formatting of the rules. To ease understanding of the rules, the system groups related rules in a ruleset according to an algorithm we developed for the grouping of related source code fragments~\cite{baum2017icsme}. Furthermore, numeric splits points are chosen to reduce the number of digits in the decimal representation.

\subsection{Interactive Data Preparation and Cleaning}

One of our goals is to support expert feedback in the whole process, not only in the mining step in the narrow sense. An important part of the data analysis process is data preparation and cleaning. Our system allows these tasks to be performed iteratively and at any time, too. When the user notes a problem in the data, he or she can remove the respective record (or all records satisfying certain criteria) or correct the class label. The system then automatically re-evaluates all rules and updates the Pareto front if needed.

One of the central limitations of rule-based models based on simple propositional logic is that they can only define hyper-rectangles in feature space. When the user notes that this limitation impedes the creation of good rules, he or she can interactively add new features calculated based on other features\footnote{with an expression language based on JavaScript}. The mining agents will then take these new features into account.

\section{Recovering Scientific Rigor}
\label{sec:scientificRigor}

Scientific rigor demands that the results of scientific studies are traceable, i.e., others shall be able to see which data led to which results, and reproducible, i.e., others shall be able to reproduce the results. Data mining studies usually perform well on these two quality criteria, because the criteria can be easily satisfied by publishing the used data and analysis source code. This nice property breaks down as soon as human feedback enters the scene: The results of the study now depend on the human input. To recover traceability and reproducibility, the proposed data mining system creates a detailed log file. It contains information on the actions taken by the user, i.e., when, with which parameters and with which results they were performed. It also contains information on the work of the mining agents, including the seeds for the pseudo-random number generators. With this log file, the final output still depends on the subjective human input, but at least it can be analyzed and re-evaluated by independent researchers. This solution is similar to the use of detailed logs in qualitative data analysis, especially when using CAQDAS software~\cite{friese2012qualitative}.



\section{Related Work}

In the following, we discuss related work.
With our system, we aim to support the whole data mining process. A survey of various data mining and knowledge discovery process models was performed by Mariscal \etal~\cite{mariscal2010survey}. Most of these models acknowledge that the process is iterative and interactive.
In line with this, the panelists at a 2002 SIGKDD panel regarded cooperation between human and computer for data mining as beneficial~\cite{ankerst2002report}.
Holzinger~\cite{holzinger2016interactive} discusses interactive machine learning in health informatics.
Interactive data mining has been studied for example by Hellerstein \etal~\cite{hellerstein1999interactive}, Zhao \etal~\cite{zhao2007ics}, and, with a multi-objective approach, by M\"uhlbacher \etal~\cite{muhlbacher2018treepod}. M\"uhlbacher's TreePOD system emphasizes the visual exploration of a two-dimensional Pareto front of decision trees. Some approaches load off most of the construction work to the user~\cite{ankerst2000towards,han2002interactive}. More similar to our approach is ``constraint-based data mining''~\cite{han1999constraint}, in which the user restrict the search space for association rules using ``rule constraints'' and guides the search with ``interestingness constraints''.

In the current study, we assume that simpler models and models with fewer features are more comprehensible~\cite{huysmans2011empirical}. These are not the only factors that influence comprehensibility~\cite{furnkranz2018cognitive,pazzani2000learning}.
Dam \etal~\cite{dam2018explainable} state that for software analytics, explainability is as important as accuracy. Freitas~\cite{freitas2014comprehensible} suggests regarding comprehensibility as one objective in a multi-objective approach.
Other researchers use a two-step process: First, a black box model is learned, and this model is transformed into a comprehensible model~\cite{johansson2004accuracy,zhang2005drc,moeyersoms2015comprehensible}. Still another approach is to create explanations for the model's decision for a specific instance upon request~\cite{tan2015online,dam2018explainable,ribeiro2016should}. However, these attempts on posthoc explainability are not without criticism, due to the associated risks~\cite{rudin2018please}.

We use a variant of multi-objective GRASP-PR~\cite{resende2016optimization,marti2015multiobjective} for rule mining. So did Ishida \etal~\cite{ishida2008exploring,ishida2009multiobjective}, Reynolds \etal~\cite{reynolds2009multi} and Pavanelli \etal~\cite{pavanelli2014extraction}, with promising results. All four studies differ from ours in the specific meta-heuristic operators and various other details and do not take domain expert feedback into account. Many other studies use multi-objective evolutionary algorithms for data mining~\cite{dehuri2006predictive,dehuri2008application,kaya2010autonomous}.
The literature on data mining with meta-heuristics is vast, so we cannot survey it exhaustively here. Early studies were done by De Jong \etal~\cite{de1993using} and Janikow~\cite{janikow1993knowledge}, with successors for example by Eggermont \etal~\cite{eggermont1999comparison}, Fidelis \etal~\cite{fidelis2000discovering}, Bernard\'o-Mansilla \etal~\cite{bernado2005domain}, and Baykaso\u{g}lu \etal~\cite{baykasouglu2007mepar}. Kwedlo \etal~\cite{kwedlo2001evolutionary} explicitly discuss cost-sensitivity in their approach.
Freitas~\cite{freitas2003survey} surveys further applications of evolutionary algorithms for data mining.

The rule generation in our system is derived from standard rule mining algorithms~\cite{breiman2001random,lavravc2010explicit,furnkranz2015brief}. The generated candidates are optimized with heuristic search. A similar approach was used by Ryan \etal~\cite{ryan1998evolution}, who hybridized C4.5 and genetic programming to improve scalability. Hybridization of evolutionary algorithms is studied more deeply by Grosan \etal~\cite{grosan2007hybrid}. Another influence for our approach was the ``shooting'' procedure for multi-objective optimization~\cite{benson1997towards,wagner2007pareto}. Further inspiration was ROCCER's approach of constructing a convex hull in ROC space~\cite{prati2005roccer}.

We apply our mining approach for prediction in software engineering. Over the last two decades, defect prediction and other applications of data mining in software engineering have become vast research areas. Literature surveys \cite{hall2012systematic,Radjenovic2013,malhotra2015systematic} provide an overview of this area.

\section{Future Work}

The current implementation of our system~\cite{baum2018predictionOnlineMaterials} is application-specific, but many of the concepts are not. Furthermore, the concepts are separable to some degree: The mining algorithm is independent of the system's exploration capabilities; the algorithm does not depend on the specifics of the evaluation function (albeit some design choices could have been different if the evaluation of a ruleset were less time-consuming); and in the overall approach as depicted in Figure~\ref{fig:approachOverview}, it would even be possible to replace rulesets with other human understandable models that allow fine-grained feedback. By adjusting the implementation's architecture to reflect this separability, the approach would be more easily reusable by other researchers. It could even be possible to devise adapters to data mining suites like Weka~\cite{dataMining2011}, which already contains simple user-driven classifiers.

In its current form, the algorithm can be used for binary classification. Future work should extend it for multi-class problems. The algorithm contains multiple options for tweaking, like the probability of generation versus perfection in the agent's main loop, or the strategy for choosing rulesets to optimize and combine. So far, we only studied these options unsystematically for our application area. Future work could try to systematically determine which variants are best under which conditions.

As noted in the previous section, the explainability of opaque models, like deep neural networks, is heavily researched at the moment. One approach is to build a simplified human-understandable model that approximates the original model. As the right balance between understandability and approximation quality is of prime importance here, a multi-objective user-informed approach like ours could be beneficial.

\section{Conclusion}

In our software engineering research, we faced two problems that we could not solve to our satisfaction with current approaches. Both share certain properties: (1)~The evaluation of model quality is domain-specific, computationally expensive compared to simple classification problems, and depends on the association of records to `cause objects'. (2)~The problems are cost-sensitive and benefit from viewing them as multi-objective problems. (3)~Acceptance by domain experts is essential, and their feedback shall be integrated. Details on one of these applications are available in a separate report~\cite{baum2018miningStudy}, the other study is still underway.

To tackle these problems, we devised a rule mining system and algorithm. The algorithm combines ideas from several areas, like research on meta-heuristics and rule learning. It is based on continuously working mining agents that generate candidate rules, optimize them by local search, and combine promising rules in search for better ones. This leads to a continuously improving approximation of the Pareto front. Domain experts can view the current mining results at any time, and can provide feedback that delimits the search space and guides the search. The system not only allows exploration of the solution space but also of the input data.

Comprehensible models, like the rulesets that we use, definitely have their limitations, and our approach will not suit every data mining problem. Still, we believe that many data mining problems share some properties with our application examples, and hope that our approach can benefit these areas. The full implementation of our system is available online~\cite{baum2018predictionOnlineMaterials}.

\appendix


\section*{Bibliography}
\bibliographystyle{elsarticle-num-names}
\bibliography{C:/Users/ich/Documents/literatur/literatur}

\end{document}